\documentstyle[11pt,fullname]{article}

 \twocolumn 

\title{\vspace*{-1cm}{\bf Morphology  with a Null-Interface}}

\author{{\bf Harald Trost} and {\bf Johannes Matiasek}\\
        Austrian Research Institute for Artificial Intelligence\thanks{%
          Financial support for the Austrian Research Institute for
          Artificial Intelligence is provided by the Austrian {\em
            Ministry of Science and Research\/}. We would like to
          thank Wolfgang Heinz for valuable comments and
          suggestions}\\
        Schottengasse 3, A-1010  Vienna, Austria \\
       {\normalsize \tt \{harald,john\}@ai.univie.ac.at}}

\date{}

\newcommand{\hpsg}{{\rm HPSG}}
\newcommand{\xtwomorf} {{\sc X2MorF}}

\newcommand{\codesize}{\small}
\newcommand{\avmsz}{\footnotesize}
\newlength{\typeindent}
\newcommand{\avmt}[2]{
\sbox{\boxtmp}{%
\settowidth{\typeindent}{\avmsz\it{#1}}\hspace{\typeindent}%
$\left[\mbox{\avmsz\it\begin{tabular}{@{\sc}l@{\,}l@{}}#2\\[-3ex]%
\makebox[0pt][r]{\raisebox{-2ex}[0pt][0pt]{\avmsz\it #1\
\,}}\end{tabular}}\right]$}
\begin{tabular}{@{}l@{}}
\usebox{\boxtmp}
\\ \\[-2ex]
\end{tabular}}

\newcommand{\idx}[1]{\fbox{\scriptsize #1}}
\newsavebox{\boxa}
\newsavebox{\boxb}
\newsavebox{\boxc}
\newsavebox{\boxd}
\newsavebox{\boxe}
\newsavebox{\boxtmp}

\begin{document}
\maketitle\thispagestyle{empty}
\vspace{-1cm}
\begin{abstract}
  \noindent We   present  an    integrated architecture   for
  word-level   and  sentence-level processing  in  a unification-based
  paradigm. The  core  of the system  is  a  CLP implementation  of  a
  unification  engine   for feature  structures  supporting relational
  values.   In this framework   an \hpsg-style grammar is implemented.
  Word-level   processing uses   \xtwomorf, a morphological  component
  based  on    an extended version    of  two-level  morphology.  This
  component is tightly integrated with the grammar as a relation.  The
  advantage of  this approach is that morphology  and syntax  are kept
  logically autonomous while at the   same time minimizing   interface
  problems.
\end{abstract}



\section{Introduction}
\label{sec:intro}


Over the  last  few  years  there  has   been a  growing  interest  in
computational morphology and phonology.  A number of systems have been
developed that deal with word-level processing. A widely used approach
is finite-state morphology,  most notably two-level morphology (for an
introduction, see  \namecite{Sproat92}).  Morphological components are
successfully used for a   wide range of stand-alone  applications like
spelling correction and hyphenation.   One obvious application  is the
use    in NLP systems  geared   to   the analysis/generation of  text.
Surprisingly, they have not  been widely applied in  this domain up to
now.

A major reason for this is  the problem of interfacing morphology with
syntax.   Reflecting the current trend   in syntax towards lexicalism,
unification-based systems use highly  structured feature structures as
input.  Translating the output of morphological components into such a
representation has proved to be difficult. Reducing interface problems
is therefore crucial to success.

A  tight integration between word  and  sentence level processing also
has linguistic advantages.  The boundary between morphology and syntax
is fuzzy.   When processing written  text the units morphology  has to
deal with are, in a  technical sense, not  words but character strings
separated by delimiters. While these strings roughly correspond to the
words of a  sentence there  are  problematic cases.   In German, e.g.,
{\em zu\/}-infinitive or verbs with separable  prefixes are written as
a single unit in some instances and separately in others.

The problem  has been recognized  and some possible remedies have been
proposed.  They  all  try to minimize or   to  eliminate the interface
between word   and   sentence level  processing.    One  step   is the
description of word formation in  terms of a unification-based grammar
to  make the result of morphological  processing directly available to
syntax and   vice versa, an   approach already  taken in \xtwomorf\
\cite{Trost90,Trost91}, an extension of two-level morphology.

The harder  problem is  the  integration  of morphophonology  which is
traditionally  formalized in a  way  not easily translatable into  the
feature formalism.  We will show  how this  can be achieved by merging
the word-level grammar of \xtwomorf\  into an \hpsg-style grammar, and
by adopting a relational view of its two-level rules.

In this paper we  assume basic familiarity with  unification-based NLP
techniques and two-level morphology.


\section{Integrating Morphology into \hpsg}
\label{sec:hpsg}


Head-driven    Phrase   Structure   Grammar  (\hpsg,   \namecite{PS1},
\namecite{PS2}) can be viewed as a mono-level but multi-stratal theory
of  grammar, where  different strata  relate to  different  aspects of
linguistic information,  but   are  represented uniformly  in  feature
logics. As  such  it is well   suited as a  linguistic theory  for our
enterprise.

\hpsg\ differentiates between three  strata---{\sc phon}, {\sc synsem}
and {\sc  dtrs}.  Though morphology is  not considered in the standard
approach, it suggests  itself to  be  included as a fourth  stratum by
introducing a feature  {\sc    morph} into  the type  {\em    sign\/}.
Morphotactics are  easily  described  in  terms  of a   feature  based
grammar.  The    problem is how  to  deal  with  morphophonology.  Two
proposals have been made to overcome this problem.

\namecite{Kriegeretal93} encode  finite state automata directly in the
feature formalism.   Since two-level rules  can be compiled  into such
automata, morphophonology can be straightforwardly integrated into the
grammar.  While  this is formally   elegant  it seems  to be  no  good
solution  for practical  considerations.  First,  it  is  not entirely
clear from  their  paper how the   problem of null characters   can be
handled.  Second, encoding large automata  will result in a very large
and unwieldy type   hierarchy.  In general, introducing automata  into
feature structures and encoding morphophonology directly at that level
seems to be too low-level.

\namecite{BirdKlein93} argue  against the use of  two-level morphology
because of  linguistic  considerations.  The linguistic  background of
two-level  rules---main stream segmental  phonology---has  widely been
rejected as a   valid   linguistic model.  Instead, they    base their
implementation      on   autosegmental             phonology      (cf.
\namecite{Goldsmith90}).

This is certainly linguistically appealing.  But there are reasons for
sticking to a more conservative  approach.  Finite-state morphology as
a formalism is not necessarily tied to segmental phonology.  There are
various approaches to cope  with non-concatenative phenomena---one  of
them   \xtwomorf\ \cite{Trost90}.   Also,   for a  number of languages
complete sets  of  two-level rules  do  exist  and can immediately  be
brought  to bear.  Finally, finite-state  morphology has  proven to be
efficient while the method proposed by \namecite{BirdKlein93} seems to
be computationally costly.

Like the  other approaches ours is  also  based on  \hpsg. However, we
employ  a different approach to integration.    Our grammar is encoded
using a   unification engine based   on constraint  logic  programming
(CLP).  Besides  conventional attribute-value descriptions this system
allows  for the direct  representation of  more general relations,  as
they are required  by \hpsg.  This  extension of the formalism is used
for the integration of morphology.  Thus \xtwomorf\  is treated as one
special relation of  the grammar.  As  a result, our approach is  more
modular than the others.   While being fully integrated morphology can
still be viewed as an autonomous component leading  to a more flexible
design.

We  will now  give an overview    of \xtwomorf\ before  describing the
integrated system and its implementation in detail.


\section{Word Level Processing --- \xtwomorf}
\label{sec:x2morf}


\xtwomorf\ differs from standard two-level morphology in two important
respects.  Continuation  classes are replaced  by a feature-based word
grammar.  This allows for a  more fine-grained description of  morphs.
It is also    a  prerequisite   for    a tight integration  with     a
unification-based grammar.  \xtwomorf\ uses a morph lexicon where each
morph has one  or more feature structures  assigned.  The word grammar
itself is simple.  Morphs have  a functor-argument structure along the
lines of  \namecite{SciulloWilliams87}.   Affixes are  unary  functors
while stems are arguments  without any further structure, resulting in
a binary tree structure.

The  other extension  concerns      the two-level rules,  which    are
supplemented   with a morphological filter   consisting  of a  feature
structure.   This is  important because  in  morphophonology only some
rules are purely phonologically motivated.   Others are triggered by a
mixture of phonological and morphological facts.  Such rules cannot be
properly represented in the standard approach.

Take, e.g.,  umlaut and schwa  epenthesis in German:  The third person
singular present tense suffix for German verbs is {\em -t}, e.g., {\em
  sag-t $\rightarrow$ sagt}.  For  stems ending in  a dental, schwa is
inserted  before the  ending, e.g.,  {\em bad-t $\rightarrow$  badet}.
This  rule does not hold across  the whole vocabulary though. Stems of
the  strong paradigm do exhibit  umlaut  in 3rdPersSgPres which blocks
schwa epenthesis.    The  final dental  of the   stem must  be omitted
instead, e.g., {\em rat-t $\rightarrow$ r\"at}.

The three  rules\footnote{These rules as well  as other data presented
  in  the examples are simplified   for the purpose of  demonstration}
shown in Fig. \ref{fig:mrules}---together with the appropriate entries
in the morph  lexicon (cf.  Fig.  \ref{fig:rat+t} below)---produce the
required behavior. In particular, these rules relate surface {\em
  r\"at\/} to lexical {\em \$rAt+t\$\/}\footnote{%
  The lexical  character {\em A\/} may  have  the surface realizations
  {\em  a\/}  and  {\em \"a\/}.  The  rule has   an empty phonological
  context  but a  morphological filter.  This  is an  example for  the
  treatment of non-concatenative phenomena in \xtwomorf.}.
\begin{figure*}[htb]
  \begin{center}
    \small
    \leavevmode
    \fbox{%
    \begin{minipage}[t]{14cm}
\makebox[2cm][r]{(i)} {\em \makebox[1cm]{A:\"a} $\Longleftrightarrow$
  \_ ;
  {\sc [morph$\mid$mhead$\mid$umlaut {\it aou-umlaut\/}]}} \\
\makebox[2cm][r]{(ii)} {\em \makebox[1cm]{t:0} $\Longleftrightarrow$
  \_   +:0  t}  \\
\makebox[2cm][r]{(iii)} {\em \makebox[1cm]{+:e}  $\Longleftrightarrow$
  dental  \_ +:0 $ [s \mid t]$   ;
  {\sc  [morph$\mid$mhead$\mid$epenthese +]}}
    \end{minipage}
    }
  \vspace{-3ex}
  \end{center}
  \caption{Three extended two-level Rules}%
  \label{fig:mrules}
\end{figure*}
\xtwomorf\  can be seen  as a relation between   a surface string (the
word  form), a  lexical    string,  and  a feature structure      (the
interpretation   of   the word  form).    Relevant for  sentence level
processing  is the morphosyntactic information  and the stem, found as
the values of paths {\sc  morph$\mid$mhead} and {\sc morph$\mid$stem}
respectively (cf. Fig.  \ref{fig:result} below).  This is supplemented
by lexeme specific  information in the value of   {\sc synsem} (for  a
detailed description see \namecite{Trost93}).


\section{Implementing \hpsg\ in a CLP Framework}
\label{sec:CLP}


\hpsg\  employs  strongly typed    feature  structures together   with
principles  constraining    them  further.     {\em  Well-typedness\/}
requirements  restrict the space   of  valid feature structures   (cf.
\namecite{Carpenter92}):  Every  feature structure must  be associated
with a type, and every type restricts its associated feature structure
in  that only  certain features are   allowed and the values of  these
features must  be  of   a certain  type.  Appropriateness    and value
restrictions are inherited along the type hierarchy.

\begin{figure}[htb]
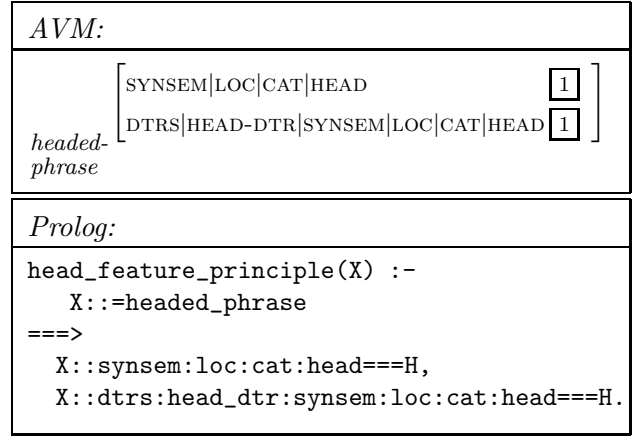

  \begin{center}
  \vspace{-2ex}
    \leavevmode
    \begin{tabular}{|l|}
      \hline
      \\[-2ex]
       {\em AVM:} \\
      \hline
      \begin{minipage}{78mm}
      \vspace*{1ex}
      \avmt{\begin{minipage}[t]{1.1cm}
       headed-\\[-0.5ex]
       phrase
      \end{minipage}}{
        synsem$|$loc$|$cat$|$head & \idx{1} \\ \\[-2ex]
        dtrs$|$head-dtr$|$synsem$|$loc$|$cat$|$head & \idx{1}\hspace{0.5em} }
      \\[3ex]
      \end{minipage}
      \\
      \hline \hline
      \\[-2ex]
      {\em Prolog:} \\
      \hline
      \begin{minipage}{78mm}
        \vspace*{1ex}
        \codesize
        \begin{verbatim}
head_feature_principle(X) :-
   X::=headed_phrase
===>
  X::synsem:loc:cat:head===H,
  X::dtrs:head_dtr:synsem:loc:cat:head===H.
        \end{verbatim}
        \vspace{-3ex}
      \end{minipage}
      \\ \hline
    \end{tabular}
    \vspace{-4ex}
  \end{center}
  \caption{Head Feature Principle}
  \vspace{-3ex}
  \label{fig:hfp}
\end{figure}

The   second   source   of  constraints,  in    order   to  admit only
linguistically    valid feature  structures,   are  the  principles of
grammar.   \namecite{PS1}   allow  general  implicative   and negative
constraints   in the form    of  conditional feature   structures.  In
\namecite{PS2} principles are given  only in verbal form.  Recent work
on formalizing the basis of \hpsg\ models them as constraints attached
to types (e.g., \namecite{CPF91}).  However, these distinctions affect
only how   the   applicability of a   principle  is  specified.   More
important for  our present purpose  is the  form which the constraints
expressed  by a principle    may take.  Besides  constraints enforcing
simple structure  sharing (e.g., the Head  Feature Principle  given in
Fig.\ref{fig:hfp})  there are   also complex  relational  dependencies
(e.g.,  in   the  Subcategorization Principle\footnote{``In   a headed
  phrase (i.e., a phrasal sign whose {\sc dtrs}  value is of sort {\em
    head-struc\/}), the {\sc subcat} value of the head daughter is the
  concatenation of  the phrase's {\sc  subcat} list with the  list (in
  order of  increasing  obliqueness)  of  {\sc synsem} values   of the
  complement   daughters.''\cite{PS2}}).     Constraints like these go
beyond the  expressivity of pure feature formalisms  alone and need to
be defined in a recursive manner.

In   order  to integrate  such  complex  constraints  in  the  feature
unification framework  we   interpret  unification of   typed  feature
structures under  the    restrictions of  principled  constraints   as
constraint  solving  in   the CLP  paradigm  \cite{clp}.

In   CLP the notion  of unification  is   replaced by the more general
notion of constraint solving. Constraint solvers  may be embedded into
a logic programming language either by writing a meta-interpreter or by
making   use of  a  system  which   allows  for  the implementation of
unification extensions.

The second approach  is  taken by DMCAI  CLP\footnote{DMCAI CLP  is an
  enhanced version of SICStus  Prolog, available by anonymous ftp from
  {\tt ftp.ai.univie.ac.at}}  \cite{Holzbaur92}, a  Prolog    system
whose unification  mechanism is extended in  such a way  that the user
may introduce interpreted terms and specify  their meaning with regard
to  unification  through Prolog predicates.   The  basic  mechanism to
achieve this behavior is   the  use of {\em  attributed  variables\/},
which  may be    qualified   by arbitrary    user-defined  attributes.
Attributed variables  behave  like ordinary Prolog  variables with two
notable exceptions: when an attributed  variable is to be unified with
a   non-variable term or another  attributed  variable the unification
extensions come into play.   For either case  the user has to supply a
predicate which explicitly specifies  how the attributes interact  and
how they should  be interpreted with respect  to the semantics of  the
application  domain.   Unification succeeds  only  if these constraint
solving  clauses managing  the  combination---or verification---of the
involved attributes are successful.

The implementation of typed feature structures in our system makes use
of  the CLP  facilities provided   by  this  enhanced Prolog   system.
Feature  structures  are  implemented   by   the  attribute {\small\tt
  fs(Type,Dag,Goals)},   where    {\small\tt   Dag}  is  a   list   of
feature-value  pairs (which is  empty  in case of   atomic types) or a
marker indicating uninstantiatedness   of the  substructure   (feature
structures are instantiated  lazily).  {\small\tt Goals}  is a list of
delayed  constraints (see   below).   Well-typed   unification of  two
feature  structures is implemented  via the constraint solving clauses
mentioned  above, taking   into  account type   hierarchy and  feature
appropriateness      (for        a      detailed   description     cf.
\namecite{MatiasekHeinz93}).

Constraints  imposed onto  feature structures   by  the principles  of
grammar are  stated  in  a  conditional form where  the  antecedent is
restricted to  contain only typing requirements.\footnote{This is only
  a syntactic   variant   of attaching  constraints  solely  to  types
  \cite{CPF91} and does not  permit general conditional  structures as
  used in  \namecite{PS1}.} In order  to account for these conditional
constraints  we  adopt  a  licensing  view:  Every node  of  a feature
structure has to be licensed by all principles of grammar.

A node  is  {\bf licensed} by a   principle if either  {\em (i)\/} the
feature structure  $F$  rooted  in  that  node  {\em satisfies\/}  the
applicability conditions   of  the   principle and  the    constraints
expressed  by   the principle  successfully unify  with  $F$,  or {\em
  (ii)\/} the feature structure   $F$  rooted in   that node is   {\em
  incompatible\/} with the  applicability conditions of the principle.
The interesting  case arises  when a  feature  structure does {\em not
  satisfy} the  applicability conditions of  the principle but is {\em
  compatible} with them.   Thus applicability of  the principle can be
decided only  later, after further  instantiation or unification steps
have restricted the (sub)structure rooted at  that node.  In precisely
this case the application (or the abandoning) of the constraint has to
be delayed.  The  delay mechanism utilizes  the {\small\tt Goals} slot
in the {\tt fs/3}\footnote{{\tt pred/n}   is the usual notation for  a
  n-ary  Prolog predicate.} attribute,  which is dedicated to hold the
delayed constraints.   As an example take the  well known Head Feature
Principle  of  \hpsg\ (Fig.\ref{fig:hfp})\footnote{The   operators {\tt
    ::=,   ::,  :, ===}    are defined  for typing    of a node,  path
  restriction,  path  concatenation and  value  restriction (type {\em
    or\/} coreference) respectively.}.
The conditional  operator {\tt ===>}   is translated at read  time via
{\small\tt  term\_expansion/2} and  implements the  delay mechanism by
compiling precondition   checks into the  principle.  These antecedent
checks trigger  either   the  application   of  the principle,     its
abandonment,  or its delay (by  annotating the variables which are not
sufficiently constrained to decide  on the antecedent with the delayed
goals).

\begin{figure}[htb]
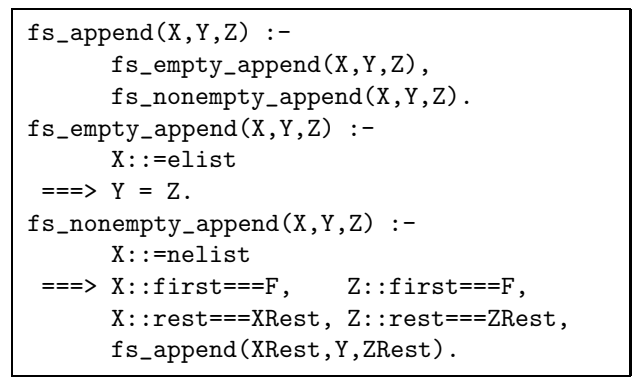

\begin{center}
  \vspace{-2ex}
  \begin{tabular}{|c|}
    \hline
  \begin{minipage}{78mm}
    \vspace*{1ex}
    \codesize
    \begin{verbatim}
fs_append(X,Y,Z) :-
      fs_empty_append(X,Y,Z),
      fs_nonempty_append(X,Y,Z).
fs_empty_append(X,Y,Z) :-
      X::=elist
 ===> Y = Z.
fs_nonempty_append(X,Y,Z) :-
      X::=nelist
 ===> X::first===F,    Z::first===F,
      X::rest===XRest, Z::rest===ZRest,
      fs_append(XRest,Y,ZRest).
    \end{verbatim}
    \vspace{-4ex}
  \end{minipage}
  \\  \hline
  \end{tabular}
  \vspace{-4ex}
\end{center}
\caption{Append for feature structure lists}
\label{fig:fsapp}
\vspace{-1ex}
\end{figure}

Two advantages  of  this approach to  implement principled constraints
are  especially important for   our  present purpose:  First,  stating
redundant  typing requirements for  embedded  structures  (i.e.   type
restrictions that would follow  automatically from well-typing) forces
delay of the  conditional  constraint  until these substructures   are
instantiated.   This  device  can, e.g.,   be used  to  block infinite
recursion in recursively defined  constraints.  Second, the right hand
part of    the   conditional is not   restricted    to feature logical
expressions, but instead can contain  arbitrary Prolog goals.  In this
way  constraints involving    relational  dependencies (such   as  the
Subcategorization Principle and  the morphological  relation between a
lexical  and a  surface string)  can  be expressed within the  feature
formalism and there  is no need for  external devices controlling this
interaction.  Furthermore, the  conditional  constraint syntax is  not
restricted  to  unary licensing principles  but can   also be  used to
express  relations, such   as {\small\tt   fs\_append/3}---needed  for
implementing   the Subcat    Principle---which  appends two    feature
structure lists (Fig. \ref{fig:fsapp}).
%
%
\begin{figure*}[t]
  \begin{center}
  \vspace{-2ex}
    \leavevmode
    \mbox{}\hspace{-0.1in}
    \begin{tabular}{|l|c|}
      \hline
      \makebox[0.82in]{}&
     \rule[-2mm]{0pt}{6mm}{\em  t:0 $\Longleftrightarrow$ \_ +:0 t \/}\\
      \hline
      {\small\em Input\/} &
      \rule[-2mm]{0pt}{6mm} {\small\tt \_ <=>   t:0   <=> ['+':0, t:t].}\\
      \hline
      {\small\em Compiled\/} &
      \begin{minipage}{5in}
        \footnotesize\tt\vspace*{1ex}
         morphrule([116,43,116|LS],[Sc,48,116|SS0],SS,LCon,SCon,F) :-\\
          \hspace*{8mm}!, Sc=48, \\

\hspace*{8mm}morphology([43,116|LS],[48,116|SS0],SS,[116|LCon],[H|SCon],F).
          \vspace{1ex}
      \end{minipage} \\
      \hline
    \end{tabular}
  \end{center}
  \vspace{-3ex}
  \caption{Sample Two-Level Rule}
  \label{fig:twolr}
\end{figure*}
\begin{figure*}[t]
  \codesize
\begin{center}
  \vspace{-2ex}
  \begin{tabular}{|c|}
    \hline \\[-2ex]
  \begin{minipage}{6in}
    \begin{verbatim}
morphology(LexStream,SurfStream0,SurfPlainIn,LexContext,SurfContext,F) :-
  instantiate(LexStream,SurfStream0,SurfPlainIn,SurfPlainOut,F),
  morphrule(LexStream,SurfStream,SurfPlainOut,LexContext,SurfContext,F).
    \end{verbatim}
    \vspace{-4ex}
  \end{minipage}
  \\ \hline  \\[-2ex]
  \begin{minipage}{6in}
    \begin{verbatim}
instantiate([LC|LCs],[SC|SCs],SurfPlainIn,SurfPlainOut,F) :-
  valid_alphabet_pair(LC,SC],
  synchronize([SC|SCs],SurfPlainIn,SurfPlainOut),
  lookahead(LC,LCs,SCs,SurfPlainOut).
    \end{verbatim}
    \vspace{-4ex}
  \end{minipage}
  \\ \hline \\[-2ex]
  \begin{minipage}{6in}
    \begin{verbatim}
synchronize([48|_],Stream,Stream) :- !.
synchronize([Char|_],[Char|Stream],Stream).
    \end{verbatim}
    \vspace{-4ex}
  \end{minipage}
  \\ \hline
  \end{tabular}
  \vspace{-3ex}
\end{center}
\caption{The {\tt morphology} relation}
\vspace{-3mm}
\label{fig:morph}
\end{figure*}
%
%
Note  that  disjunctive relations such  as {\em  append\/} can  now be
written    as  the  conjunction of    two   specialized cases applying
conditionally.   Furthermore,  infinite  loops   due to uninstantiated
variables can  never occur,  a  crucial  requirement when  integrating
relational dependencies into a lazy instantiating feature formalism.


\section{Embedding \xtwomorf\ into the Feature System}
\label{sec:system}


Originally \xtwomorf\  was   realized  as  a   separate  morphological
component interfaced  to  the  sentence analyzer/generator  only   via
sequential   data transfer.   In  the case  of   analysis, the feature
structure representing  the word form was   transmitted to the parser.
For generation, \xtwomorf\  expected  a  feature structure as    input
reproducing   one or  more   word    forms.  This  purely   sequential
architecture was not satisfactory because of the problems mentioned in
the introduction.

In order to  achieve tight integration,  we adopt a relational view of
\xtwomorf\ and encode the relation  between surface string and lexical
string directly  without using  finite  state automata  (for arguments
supporting  this  approach  cf.  \namecite{Abramson92}).  However, our
approach    extends \namecite{Abramson92} in     that  it {\em  (i)\/}
explicitly accounts  for the  insertion  of null  characters and  {\em
  (ii)\/} introduces   the  filter  concept  of   \xtwomorf\  into the
relational approach.

The general format of a two-level rule specification in our system is
\begin{center}
  \vspace{-1ex}
  {\small\tt LCon <=> Transition <=> RCon}
{\it [\/}{\small\tt :- Filter}{\it ]\/}
\end{center}
\vspace{-1ex}
in the case  of equivalence  rules,  optional rules are written  using
only single arrows ({\tt =>} and  {\tt <=}).  These rules are compiled
into Prolog  clauses\footnote{Note   that left contexts   are encoded
  reversed to account for  the left to right traversal  of the pair of
  character streams. Left contexts can be remembered and checked most
  efficiently this  way.} relating the  lexical and  surface character
streams appropriately (see  Fig.\ref{fig:twolr} for an example of  the
{\em t-elision\/} rule for German).

To obtain a correct  relationship  between surface and  lexical string
every  transition  has  to  be  licensed   by a   morphological  rule.
Transitions  not mentioned  by  rules are  handled  by a default rule.
Instantiation of contexts may not   be done by the rules   themselves,
since  this   would make it  impossible   to obtain negation   via the
cut-operator. Instead,   it is handled  separately in  a backtrackable
fashion.

The central  relation is  the {\small\tt  morphology}  predicate, (see
Fig.     \ref{fig:morph}) mediating  between   lexical string, surface
string (with inserted null  elements), the pure (denullified)  surface
string  and the  feature   structure   of  the morphological     sign.
Instantiation of   pairs  is done  depending  on  the possible lexical
continuations (the lexicon being represented by a trie-structure). The
amount of lookahead is determined  by the current pair  which is to be
licensed  by {\small\tt morphrule}.\footnote{This  interaction and the
  lexicon lookup of the feature structure corresponding to the current
  morph, which takes  place when encountering  a morph boundary is not
  shown for  the sake of simplicity.}  Synchronization  of surface and
lexical string by insertion of null characters  is also handled at the
instantiation level.

The  integration of the  two-level relation into the general framework
of  the  feature based sentence-level  and  word-level grammars is now
performed by  adding this relation  as a principled constraint  at the
appropriate level.

In a definite   clause style AVM  notation this  could  be  written as
follows (given that {\small\tt  morphology/3} is a wrapper  around the
morphology relation given above,  starting with empty left context and
hiding the nullified surface stream):\\
\\
  \avmt{word}{
    phon & \idx{2}string \\
    morph & \idx{3}\hspace{-7mm}\avmt{msign}{
      mstring & \idx{1}string \\
      stem    & string \\
      mhead   & mhead
      }\\ \\[-2ex]
    head & head \\
    synsem & synsem
    }  \\
  \mbox{}\hfill$\longleftarrow$
  {\small \em morphology\/}(\,\idx{1},\,\idx{2},\,\idx{3}\,) \\
\\
The actual implementation as a principled  constraint in our formalism
additionally takes  care of delaying   the actual enforcement of  this
relation in case the strings are not sufficiently instantiated.

A second provision has to be made in the word level grammar to assure
proper concatenation of the lexical strings of the morphological signs
being combined. Given the subtyping of {\em msign\/} into {\em marg\/}
and {\em mfunctor\/}, which in turn has the subtypes {\em
  leftfunctor\/} and {\em rightfunctor\/}, the principled constraints
ensuring concatenation of a left functor with its argument are shown
in Fig. \ref{fig:concat}.
\begin{figure}[htb]
\begin{center}
  \vspace{-2ex}
  \begin{tabular}{|c|}
    \hline \\[-2ex]
  \begin{minipage}{78mm}
    \codesize
    \begin{verbatim}
concat_right_functor(X) :-
        X::=rightfunctor,
        X::arg:mstring===subtype_of(string)
  ===>
        X::arg:mstring===Arg,
        X::affix===Suffix,
        X::mstring===Mstring,
        concat(Arg,Suffix,Mstring).
    \end{verbatim}
  \vspace{-4ex}
  \end{minipage}
  \\  \hline
  \end{tabular}
  \vspace{-3ex}
\end{center}
\caption{Concatenation of lexical strings}
\label{fig:concat}
\end{figure}
Concatenation is delayed until the argument's {\sc mstring} is
instantiated. Thus, infinite loops when concatenating are avoided.

\begin{figure}[htb]
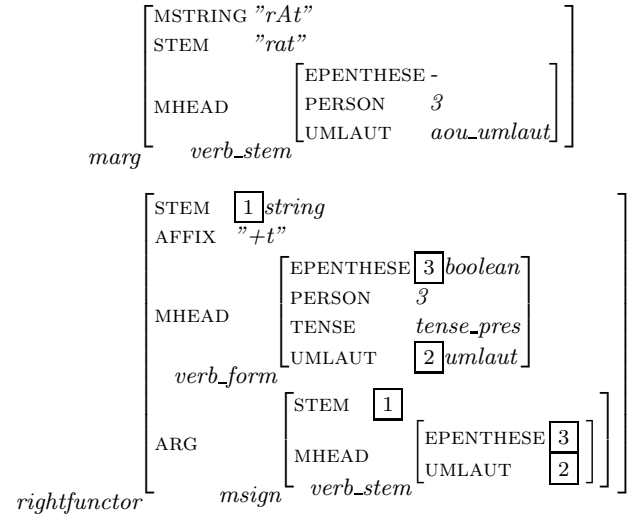

  \begin{center}
  \vspace{-2ex}
    \leavevmode
    \avmt{marg}{
      mstring & "rAt" \\
      stem &    "rat" \\
      mhead &   \hspace{-11mm}\avmt{verb\_stem}{
        epenthese &- \\
        person &   3\\
        umlaut &   aou\_umlaut}}
\\[2ex]
    \avmt{rightfunctor}{
      stem & \idx{1}string \\
      affix &"+t" \\
      mhead & \hspace{-11mm}\avmt{verb\_form}{
        epenthese &\idx{3}boolean \\
        person &   3 \\
        tense &    tense\_pres \\
        umlaut &   \idx{2}umlaut } \\ \\[-2ex]
      arg &  \hspace{-5mm}\avmt{msign}{ 
        stem & \idx{1} \\
        mhead & \hspace{-11mm}\avmt{verb\_stem}{
          epenthese &\idx{3}\hspace{1mm}\mbox{} \\
          umlaut &   \idx{2}}}}
    \vspace{-3mm}
  \end{center}
  \caption{Lexical entries}
  \label{fig:rat+t}
\end{figure}

As an example we demonstrate how these constraints interact in forming
the third person  singular present tense form of  the German verb {\em
  raten\/} (to guess). The lexical string is composed of the stem {\em
  rAt\/} and the  suffix {\em +t\/}. The lexical
entries of these two morphs are given in Fig. \ref{fig:rat+t}.

The two-level rules applicable for this example are the {\em
  t-elision\/} rule (Fig.\ref{fig:twolr}) and two rules with filters
licensing {\em a-umlaut\/} and {\em epenthesis\/}, given in the input
notation for our system (Fig.\ref{fig:frule}).
\begin{figure*}[htb]
  \begin{center}
  \vspace{-2ex}
    \leavevmode
    \begin{tabular}{|l|c|}
      \hline
      {\small\em A-umlaut}
      &
     \rule[-2mm]{0pt}{7mm} {\small\tt
        \_ <=> A:"a   <=> \_ :- filter(X, [X::mhead:umlaut===aou\_umlaut])} \\
      \hline
      {\small\em Epenthesis\/} &
      \rule[-2mm]{0pt}{7mm} {\small\tt
        dental <=> '+':e   <=> s\_or\_t :- filter(X,
[X::mhead:epenthese==='+'])} \\
      \hline
    \end{tabular}
    \vspace{-5mm}
  \end{center}
  \caption{Filter Rules}
  \vspace{-3mm}
  \label{fig:frule}
\end{figure*}

Interaction between  syntactic and morphological processes takes place
at the  word level.  The  application of  the two-level rules relating
the surface string (i.e the {\sc phon}-value of  the {\em word\/}) and
the lexical-string (i.e.  {\sc  morph$\mid$mstring}) is also triggered
here.  This  interaction is  completely   neutral with  respect to the
direction of  processing  due to   its relational  nature. Parsing  is
performed by simply instantiating the {\sc phon} value. Generation can
be achieved when {\sc morph$\mid$mstring} is present, which in turn is
obtained by concatenating the   lexical strings of the {\em  msigns\/}
instantiated by the morph grammar.

As a result of this constraint interaction the structure shown in Fig.
\ref{fig:result}  is  obtained.  Features  relevant at  the  syntactic
level (such as {\sc person} and {\sc tense})  are percolated from {\sc
  morph$\mid$mhead}  to {\sc  synsem$\mid$loc$\mid$cat$\mid$head}  via
structure sharing constraints attached  to the type {\em word\/} (this
interaction is not  shown in Fig.  \ref{fig:result}).  Information  on
subcategorization and  semantic content for the  word is obtained from
the lexeme lexicon  using    {\sc morph$\mid$stem}  as a  key.   These
constraints  complete   the    interaction    between  syntactic   and
morphological processing at the word-level.

\sbox{\boxa}{
\avmt{marg }{
  mstring &"rAt"   \\
  stem &   \idx{1}   \\ \\[-2ex]
  mhead & \hspace{-11mm}\avmt{\begin{minipage}[t]{0.75cm}
         verb-\\[-0.5ex]
         stem
      \end{minipage}}{
    epenthese &\idx{3}   \\
    person &   3   \\
    umlaut &   \idx{2}\   }\\[2ex]}}

\begin{figure}[t]
  \leavevmode
\hspace{-5mm}
  \avmt{word}{
phon & "r\"at"   \\
morph & \hspace{-13mm}\avmt{\begin{minipage}[t]{1.1cm}
       \mbox{}\hfill right-\\[-0.5ex]
       \mbox{}\hfill functor
      \end{minipage}}{
       mstring &"rAt+t"   \\ \\[-2ex]
       stem &   \idx{1}"rat" \\
       affix &  "+t"   \\
       mhead &  \hspace{-11mm}\avmt{\begin{minipage}[t]{0.75cm}
         verb-\\[-0.5ex]
         form
      \end{minipage}}{
                epenthese &\idx{3}-   \\
                person &   3   \\
                tense &    tense\_pres   \\
                umlaut &   \idx{2}aou\_umlaut}   \\ \\[2ex]
       arg &   \hspace{-13mm}\usebox{\boxa} \\}\\[2ex]}
\caption{Result of constraint interaction}
\vspace{-3mm}\
\label{fig:result}
\end{figure}


\section{Conclusion}
\label{sec:concl}


We have presented a framework for  the tight integration of word level
and sentence  level processing in a   unification-based paradigm.  The
system is   built upon a   unification engine   implemented  in a  CLP
language supporting types and definite   relations as part of  feature
descriptions. Using   this  extended   feature  formalism, which    is
independently motivated   by requirements    of  standard \hpsg,     a
reimplementation of \xtwomorf\ was  integrated  into the grammar as  a
specialized relation.

This architecture has computational as  well as linguistic advantages.
Integrating morphology and  morphophonology directly into  the grammar
is in the  spirit of \hpsg, which  views grammar as a relation between
the phonological    (or  graphemic) form  of   an  utterance   and its
syntactic/semantic   representation.    This  way  the   treatment  of
phenomena transcending  the boundary between  morphology and syntax is
also made possible.

On the implementation side, the practical problems of interfacing two
inherently different modules are eliminated. For applications this
means that using a morphological component is made easy. Nevertheless,
this tight integration still leaves morphology and syntax/semantics as
autonomous components, enabling direct use of existing data sets
describing morphophonology in terms of the two-level paradigm.


\begin{thebibliography}{}
\setlength{\itemsep}{0pt}
\small
\bibitem[\protect\citename{Abramson}92]{Abramson92}
  Abramson  H.:  A Logic  Programming View  of Relational Morphology, in
  Proceedings of the 15th COLING, August 23-28, 1992, Vol.III,
  pp.850-854, 1992.
\bibitem[\protect\citename{Bird \& Klein}93]{BirdKlein93}
  Bird S., Klein E.:  Enriching HPSG Phonology, University of Edinburgh,
  UK, Research Paper EUCCS/RP-56, 1993.
\bibitem[\protect\citename{Carpenter et al.}91]{CPF91}
  Carpenter B., Pollard C., Franz A.: The Specification and Implementation of
  Constraint-Based Unification Grammars, Proceedings of 2$^{nd}$ IWPT,
  Cancun, Mexico, 143-153, 1991.
\bibitem[\protect\citename{Carpenter}92]{Carpenter92}
  Carpenter B.: {\em The Logic of Typed Feature Structures}, Cambridge
  University Press, Cambridge Tracts in Theoretical Computer Science 32, 1992.
\bibitem[\protect\citename{Goldsmith}90]{Goldsmith90}
  Goldsmith J.A.: {\em Autosegmental and Metrical Phonology}, Basil
  Blackwell, Oxford, 1990.
\bibitem[\protect\citename{Holzbaur}92]{Holzbaur92}
  Holzbaur C.: Metastructures vs. Attributed Variables in the Context of
  Extensible Unification, in Bruyn\-ooghe M. and Wirsing M.(eds.), Programming
  Language Implementation and Logic Programming, Springer, LNCS 631,
pp.260-268,
  1992.
\bibitem[\protect\citename{Jaffar \& Lassez}87]{clp}
  Jaffar J., Lassez J.L.: Constraint Logic Programming, in Proceedings 14th ACM
  POPL Conf., Munich, 1987.
\bibitem[\protect\citename{Krieger et al.}93]{Kriegeretal93}
  Krieger H.-U., Pirker H., Nerbonne J.: Feature-based Allomorphy, Proceedings
  of the 31st Annual Meeting of the ACL,  Columbus, Ohio, pp.140-147, 1993.
\bibitem[\protect\citename{Matiasek \& Heinz}93]{MatiasekHeinz93}
  Matiasek J., Heinz W.: A CLP Based Approach to HPSG, \"Osterreichisches
  Forschungsinstitut f\"ur Artificial Intelligence, Wien, TR-93-26, 1993.
\bibitem[\protect\citename{Pollard \& Sag}87]{PS1}
  Pollard C.J., Sag I.A.: {\em Information-Based Syntax and Semantics},
  University of Chicago Press, Chicago, 1987.
\bibitem[\protect\citename{Pollard \& Sag}94]{PS2}
Pollard,  C.J,  Sag I.A.: {\em  Head-Driven Phrase Structure Grammar},
University of Chicago Press and CSLI Publications, in press.
\bibitem[\protect\citename{di Sciullo \& Williams}87]{SciulloWilliams87}
  di Sciullo A.-M., Williams E.: {\em On the Definition of Word}, MIT Press,
  Cambridge, MA, 1987.
\bibitem[\protect\citename{Sproat}92]{Sproat92}
  Sproat R.: {\em Morphology and Computation}, MIT Press, Cambridge, MA,
ACL-MIT
  Series in NLP, 1992.
\bibitem[\protect\citename{Trost}90]{Trost90}
  Trost H.: The Application of Two-Level Morphology to Non-Concatenative German
  Morphology, in Karlgren H.(ed.), Proceedings of the 13th COLING,
  Helsinki, Finland, pp.371-376, 1990.
\bibitem[\protect\citename{Trost}91]{Trost91}
  Trost H.: X2MORF: A Morphological Component Based on Augmented Two-Level
  Morphology, in Proceedings of the 12th IJCAI, Morgan Kaufmann, San
  Mateo, CA, pp.1024-1030, 1991.
\bibitem[\protect\citename{Trost}93]{Trost93}
  Trost H.: Coping with Derivation in a Morphological Component, in 6th
  Conference of the European Chapter of the ACL, Utrecht, pp.368-376, 1993.

\end{thebibliography}
\end{document}